\documentclass[letterpaper, 10 pt, conference]{ieeeconf}  % Comment this line out if you need a4paper

\IEEEoverridecommandlockouts                              % This command is only needed if 
\usepackage{graphicx}
\usepackage{setspace}
\usepackage[T1]{fontenc}
\usepackage{multicol}
\usepackage{amsmath,amssymb}
\usepackage[suffix=]{epstopdf}
\usepackage{graphicx,color}
\usepackage{xspace}
\usepackage{pifont}
\usepackage{balance}
\usepackage[per-mode=symbol,binary-units=true,range-units=single,range-phrase=-]{siunitx}
\usepackage{textcomp} % \texttimes
\usepackage{booktabs}
\usepackage{paralist}
\usepackage{footmisc}
\usepackage{comment}
\usepackage{subcaption}
\usepackage{layouts}
\usepackage{cancel}
\usepackage{url}
\usepackage{bm}
\usepackage{tikz}
\usepackage[percent]{overpic}
\usepackage{tabularx} % better tables
\usetikzlibrary{fit,shapes.callouts,shapes.geometric,backgrounds,positioning,arrows.meta}

\graphicspath{{./figures/}}

\newcommand{\eg}{e.g.,\ }

\newcommand{\etal}{\xspace{}et al.\xspace}

\newcommand{\reffig}[1]{Fig.~\ref{#1}}
\newcommand{\reftab}[1]{Tab.~\ref{#1}}
\newcommand{\refsec}[1]{Sec.~\ref{#1}}
\newcommand{\refeq}[1]{(\ref{#1})}

\title{\LARGE \bf Fast Time-optimal Avoidance of Moving Obstacles\\ for High-Speed MAV Flight}

\author{Marius Beul and Sven Behnke%
\thanks{This work has been supported by a grant of the Mohamed Bin Zayed International Robotics Challenge (MBZIRC).}% <-this % stops a space
\thanks{The authors are with the Autonomous Intelligent Systems Group, University of Bonn, Germany
        {\tt\small mbeul@ais.uni-bonn.de}}%
}

\begin{document}

\maketitle
\thispagestyle{empty}
\pagestyle{empty}

\begin{tikzpicture}[overlay, remember picture]
  \path (current page.north) ++(0.0,-1.0) node[draw = black] {Accepted for IEEE/RSJ International Conference on Intelligent Robots and Systems (IROS), Macau, China, November 2019};
\end{tikzpicture}
\vspace{-0.3cm}

%%%%%%%%%%%%%%%%%%%%%%%%%%%%%%%%%%%%%%%%%%%%%%%%%%%%%%%%%%%%%%%%%%%%%%%%%%%%%%%%
\begin{abstract}
In this work, we propose a method to efficiently compute smooth, time-optimal trajectories for micro aerial vehicles (MAVs) evading a moving obstacle.
Our approach first computes an n-dimensional trajectory from the start- to an arbitrary target state including position, velocity and acceleration.
It respects input- and state-constraints and is thus dynamically feasible.
The trajectory is then efficiently checked for collisions, exploiting the piecewise polynomial formulation. If collisions occur, viastates are inserted into the trajectory to circumvent the obstacle and still maintain time-optimality. These viastates are described by position, velocity, and acceleration.
The evaluation shows that the computational demands of the proposed method are minimal such that obstacle avoidance can begin within few milliseconds. Optimality of generated trajectories, combined with the ability for frequent online re-planning from non-hover initial conditions, make the approach well suited for evasion of suddenly perceived obstacles during fast flight.
\end{abstract}

%%%%%%%%%%%%%%%%%%%%%%%%%%%%%%%%%%%%%%%%%%%%%%%%%%%%%%%%%%%%%%%%%%%%%%%%%%%%%%%%
\section{Introduction}
\label{sec:Introduction}
In recent years, multiple novel applications for flying robots emerged, enabled by two main factors: i) manufacturers developed affordable and capable micro aerial vehicles (MAVs) for hobby, recreation, and professional usage that do not require extensive flight training; ii) recent advances in robotic research led to efficient methods for environment perception and safe navigation, making various applications that can only be performed autonomously possible.
This includes operations at high velocities and with multiple MAVs. One driver for developing such systems was also the DARPA-formulated goal of flying fast and autonomously in their Fast Lightweight Autonomy (FLA) program \cite{darpaFLA}.

In the field, applications often require fast flight in nearly obstacle-free environments. For example, in a typical outdoor inspection task, only the to-be-inspected object (windmill, power line, \dots) obstructs the otherwise free space. Also, when executing an exploration mission with multiple MAVs, often the only obstacles present are the other MAVs in mostly free space.

A possible use case for our method emerged from the Mohamed Bin Zayed International Robotics Challenges (MBZIRC) 2017 \cite{ecmr2017_c3} and 2020 \cite{MBZIRC2020}. At MBZIRC 2017, multiple MAVs shared the same workspace while picking and delivering small discs. For MBZIRC 2020, an MAV has to avoid static balloons while interacting with a flying object. Here, fast mission accomplishment is key for a high score. Since the environment is mostly obstacle free, one can omit global planning methods like A* \cite{hart_1972_sigart}. Instead, direct flight to the target, only avoiding small static no-fly zones around the balloons and dynamic no-fly zones around the other MAVs is a feasible approach.

\begin{figure}[t]
  \centering
  \includegraphics[trim=95mm 25mm 74mm 35mm,clip,width=1.0\linewidth]{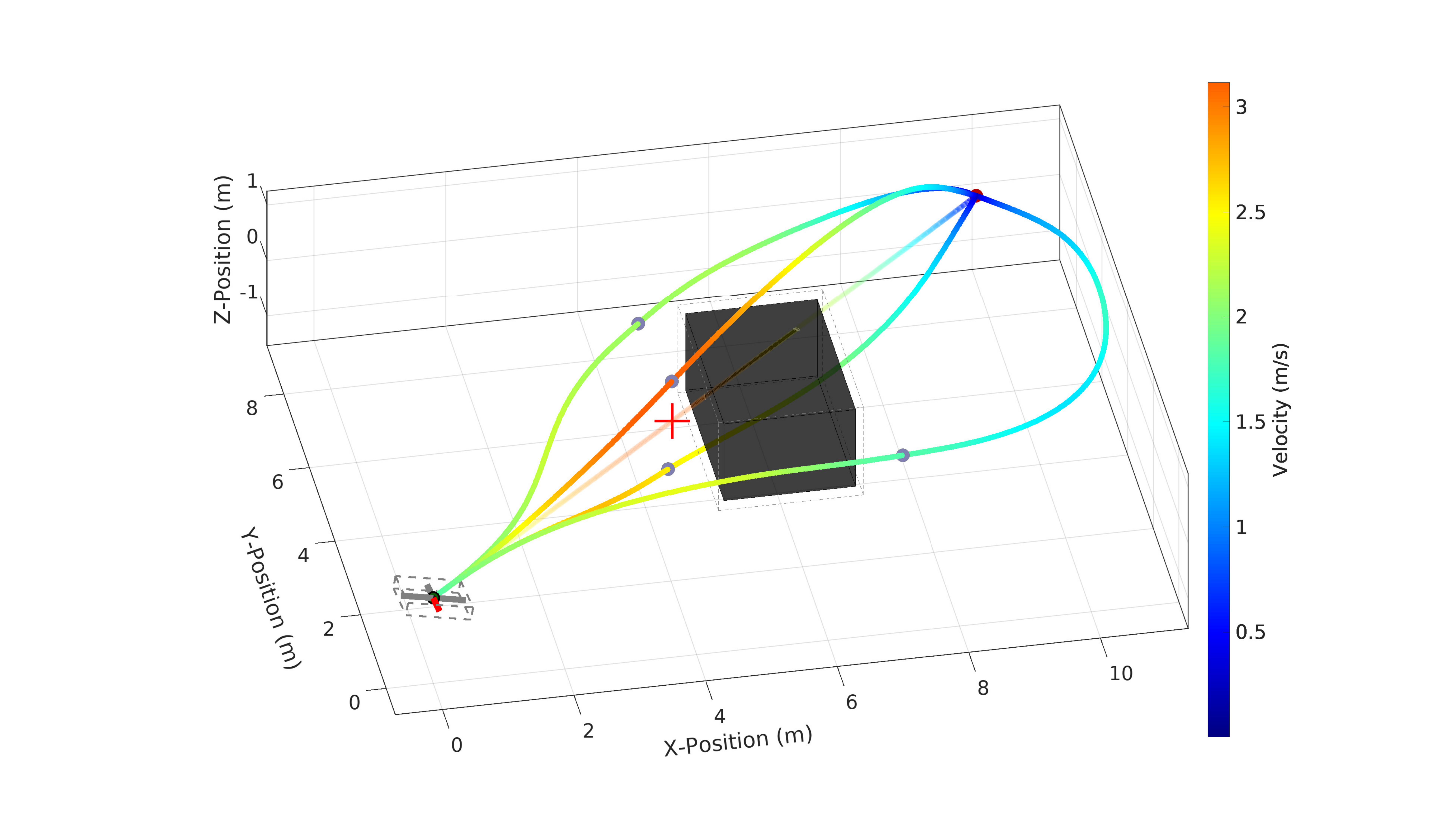}~
  %\vspace{-1ex}
  \caption{\small{The MAV on the bottom left flies towards the top right target (red dot) with a velocity of \SI{1.8}{\meter\per\second} and an acceleration of \SI{0.5}{\meter\per\second\squared} when suddenly perceiving an obstacle. Our method instantaneously detects the collision (red cross) of the original semitransparent trajectory. Within \SI{6}{\milli\second}, it computes a set of four time-optimal avoidance trajectories from the current state to the target state that steer through the gray viastates. These trajectories evade the obstacle in varying dimensions (over the top, underneath, left and right). The fastest collision-free trajectory (over the top of the obstacle) is chosen and executed by the MAV. The detour only takes \SI{7.33}{\second} vs. \SI{7.31}{\second} of the original trajectory.}}
  \label{fig:teaser}
  \vspace{-3ex}
\end{figure}

\reffig{fig:teaser} illustrates the proposed method. We first compute an time-optimal trajectory from the start state to an arbitrary target state that is efficiently checked for collisions. If collisions occur, via states described by position, velocity, and acceleration are inserted to circumvent the obstacle and maintain time-optimality. 
In our obstacle avoidance algorithm, we employ and extend methods based on our own previous work~\cite{beul2017icuas}.
 Our main contributions are:
\begin{compactitem}
    \item generation of trajectories targeting only partially defined target states (\refsec{sec:Partially_Defined_Target_States}),
    \item fast computation of optimal trajectories that avoid a static or moving obstacle (\refsec{sec:Obstacle_Avoidance}),
    \item evaluation of our method in simulation including profiling of computational requirements (\refsec{sec:Evaluation}).
\end{compactitem}

In this work, we employ our method to 3-dimensional problems for ease of explanation. The method itself however does not make any assumption about the dimensionality of the planning problem.

The code used in this work is open source\footnote{\scriptsize{\url{http://www.ais.uni-bonn.de/videos/IROS_2019_Beul}}\label{ftn:Website}}.

\section{Related Work}
\label{sec:Related_Work}
In literature, several works address collision avoidance of MAVs in dynamic environments. The approaches can be mainly categorized into by considered dimension (2D vs. 3D), complexity of the environment, runtime, and the ability to deal with non-static obstacles.\\
While planning in complex environments can be solved with sample-based planners, grid-based planners, etc., these approaches often either do not consider system-dynamics or are not fast enough for real-time planning. Hence, most obstacle avoidance techniques for MAV assume a simple environment to be able to run with the control frequency of at least \SI{10}{\hertz} instead. Since they run as a lower layer in a hierarchy of planners, the missing consideration of complex environments can be compensated by higher layers.

Zhang \etal~\cite{zhang_2018_iros} use a set of precomputed alternative 3D paths to quickly react to suddenly perceived obstacles. The set is hierarchically organized in levels such that branches from alternative paths can be efficiently stored. The method works in 3D and is real-time capable (\SI{<50}{\micro\second}) due to the precomputation. However, the precomputation needs to be performed whenever the target waypoint changes and generated trajectories are not optimal. Depending on the discretization, generated trajectories can be far from optimal whilst always collision free. Also, the evaluation only considers static obstacles.

Zhu and Alonso-Mora~\cite{zhu_2019_ral} propose a probabilistic collision avoidance method for navigation among moving obstacles. They explicitly consider the collision probability and solve a chance constrained nonlinear model predictive control problem (CCNMPC) to find valid trajectories. The method is real-time capable (average of \SI{14.3}{\milli\second} with peaks over \SI{70}{ms}) and can deal with constantly moving obstacles. The method explicitly respects uncertainty in the obstacle state. No information about the quality of the generated trajectories (\eg optimality) is given.

Gao and Shen~\cite{gao_2017_icra}, and Gao \etal~\cite{gao_2018_iros} use an optimization-based framework to generate collision free MAV trajectories. The method models both static and moving obstacles. It guarantees the trajectories to be optimal with respect to control effort, but computation is slow (\SI{1.0}{\second}).

Liu \etal~\cite{liu_2017_icra}, use safe flight corridors to plan 3D trajectories that are dynamically feasible. They use quadratic programming techniques to find trajectories in a subset of the free space defined by convex polyhedra. Replanning takes \SIrange{50}{300}{\milli\second} and is thus limitedly real-time capable.

In \cite{liu_2017_iros}, Liu \etal present a search-based motion planning method that used LQR-techniques to find dynamically feasible minimum-time trajectories in a discretized state-space. The method can generate smooth second- and third-order trajectories in 2D and 3D.
The authors claim that the method can be used for fast online re-planning of trajectories, but the runtime for the 3D third-order system---equivalent to our method---is high (avg. \SI{2.98}{\second}, max \SI{9.5}{\second}).
Also the optimality of the trajectories is governed by the discretization of the state space. Our method does not suffer from discretization-induced suboptimality. In \cite{liu_2018_icra}, Liu \etal continue the work on the search-based planning method. They now model the MAV with an ellipsoid bounding box (instead of a sphere). The characteristics of the method including its capabilities and restrictions are similar to their previous work.

Lopez and How~\cite{lopez_2017_icra} use a 3D triple-integrator MAV-model similar to ours. Like Liu, they sample the state space to find collision-free trajectories, but don´t give information on the optimality of the solution.
The authors report computation times in the low milliseconds, but due to the sampling-based nature of their method, it is unclear how the method scales with varying complexity of the environment.

Szmuk \etal~\cite{szmuk_2018_iros} use convex optimization to find 3-dimensional paths in real-time. They show avoidance of up to three moving obstacles. The optimization takes up to \SI{81.4}{\milli\second} and is thus fairly real-time capable. The authors use a double-integrator model for the problem formulation. They justify this decision by the assumption that the attitude dynamics are significantly greater than the translational dynamics. This may be true for small agile MAVs like the ones used by the authors in the evaluation, but may introduce large errors when executed on a slower MAV.

Similar to our method, Mellinger \etal~\cite{mellinger_2011_icra} present a method to calculate trajectories with free degrees of freedom (DoF). Trajectories also pass through fixed positions with optimal velocities. However, the solution is obtained by solving a constrained gradient descent problem including the numerical computation of the derivatives. In comparison to our method, generated trajectories are minimum snap and the trajectories are tracked by a separate controller instead of directly using the trajectory generator as feedforward with fast replanning.

As shown above, multiple approaches exist that generate smooth collision-free trajectories in environments with varying complexity and runtime requirements. Unfortunately, no standard benchmark exists to compare the methods, specifically for use on MAVs that require replanning times in the order of milliseconds, strict constraints on state and input variables, smoothness and satisfaction of optimality conditions like control effort or total trajectory time.

In the work of Kr\"oger~\cite{kroeger_2011_icra}, third-order polynomial trajectories are generated that are similar to our work. The method lacks any kind of obstacle avoidance, though. Nevertheless, applications that utilize this method and need obstacle avoidance can be easily transferred to our proposed approach.

To our knowledge, no method exists that can generate smooth, optimal obstacle-avoiding 3D trajectories for MAVs that can be computed within typical control-loop frequencies. The method proposed in this work replaces our reactive low-level obstacle avoidance mechanism~\cite{beul_2018_iros} that is not applicable to generating aggressive high-speed trajectories.

\begin{figure*}[t]
  \centering
  \includegraphics[trim=00mm 280mm 00mm  25mm,clip,width=1.0\linewidth]{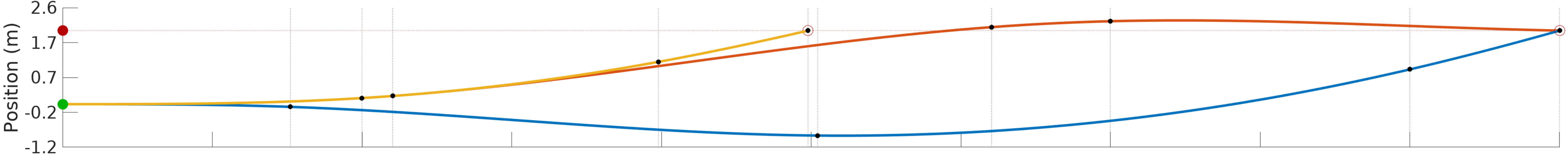}~\\
  \includegraphics[trim=00mm 200mm 00mm 105mm,clip,width=1.0\linewidth]{partially_defined.eps}~\\
  \includegraphics[trim=01mm 114mm 00mm 180mm,clip,width=1.0\linewidth]{partially_defined.eps}~\\
  \includegraphics[trim=01mm  30mm 06mm 268mm,clip,width=1.0\linewidth]{partially_defined.eps}~\\
  \vspace{-1ex}
  \caption{\small{Trajectories from state $\mathbf{x} = (0.0, 0.0, 0.0)^\intercal$ to $\mathbf{x} = (2.0, \textit{NaN}, 0.5)^\intercal$. Yellow: The undefined DoF (velocity) is chosen such that the trajectory is time-optimal. This results in a target velocity of \SI{1.9}{\meter\per\second}. Blue, Orange: Trajectories that exactly last \SI{5.0}{\second}. The undefined DoF is maximized (blue) or minimized (orange). One can see how the trajectories build up momentum by first accelerating in the opposing direction to maximize the time of acceleration (\SIrange{1.5}{5.0}{\second} for blue and \SIrange{2.1}{5.0}{\second} for red trajectory.)}}
  \label{fig:partially_defined}
  \vspace{-4ex}
\end{figure*}

\section{Partially Defined Target States}
\label{sec:Partially_Defined_Target_States}
Our method extends our previously published work~\cite{beul2017icuas} in which we generate time-optimal second- and third-order trajectories from arbitrary fully defined start states to arbitrary target states with piecewise constant jerk output.

\subsection{Previous Method for Trajectory Generation}
\label{Previous_Method_for_Trajectory_Generation}
Trajectories generated with our previous method respect per-axis constraints on minimum and maximum velocity, acceleration and jerk. Any number of individual axes can be coupled by synchronizing the total time of each trajectory. Since the method is very fast ($\ll \SI{1}{\milli\second}$ per axis per trajectory), it can be used in closed loop even for fast systems. With the ability to predict the target state, trajectories end in an optimal interception state when the target state is non-stationary. The method has been successfully used as model predictive controller on different micro aerial vehicles, in different research projects and robotic competitions.
Like most works, our previous method  requires fully defined target state inputs. Furthermore, it does not feature obstacle avoidance.

\subsection{Solving Partially Defined Target States}
\label{Solving_Partially_Defined_Target_States}
In certain situations, the targeted state cannot be defined completely, \eg when the MAV is supposed to pass through a narrow gap in a wall. Here, the 3-dimensional position and the pitch angle of the MAV is fixed such that it fits through the gap. Also the lateral velocity is fixed to zero such that the MAV passes the gap on the shortest path.
However, the velocity orthogonal to the wall can be chosen freely. Also, our obstacle avoidance method described in \refsec{sec:Obstacle_Avoidance} builds upon the ability to find trajectories for partially defined target states.

\begin{table}[b]
  \vspace{-2ex}
  \caption{\small{Combinations of possible target state variables.}}
  \small
  \centering
  \begin{tabular}{l|ccccccc}
  \toprule
  Position     & \ding{51} & \ding{51} & \ding{51} & \ding{51} & \ding{56} & \ding{56} & \ding{56}\\
  Velocity     & \ding{51} & \ding{51} & \ding{56} & \ding{56} & \ding{51} & \ding{51} & \ding{56}\\
  Acceleration & \ding{51} & \ding{56} & \ding{51} & \ding{56} & \ding{51} & \ding{56} & \ding{51}\\
  \bottomrule    
  \end{tabular}\\
  \vspace*{0.75ex}
  \scriptsize\ding{51}: the value is defined, \ding{56}: the value can be chosen freely
  \label{tab:permutations}
\end{table}

Since our pipeline assumes a 3-dimensional state for each axis of the start- and target states, the permutations stated in \reftab{tab:permutations} for the definition of target-states can occur. The first instance is the nominal condition with all derivatives defined. In the following, we describe how we treat the remaining six permutations.

In order to generate trajectories without fully defined target states, we first define the trajectory for each axis as a system of 21 differential equations:
\begin{align}
    p_{n} &= p_{n-1} + \int_{t_{n-1}}^{t_{n}}{v_{n}}\,\mathrm{d}t,\\
    v_{n} &= v_{n-1} + \int_{t_{n-1}}^{t_{n}}{a_{n}}\,\mathrm{d}t, \qquad n = \{1;\dots;7\}\\
    a_{n} &= a_{n-1} + \int_{t_{n-1}}^{t_{n}}{j_{n}}\,\mathrm{d}t,
\end{align}
with $p_{0},v_{0},a_{0}$ being the current state and $p_{7},v_{7},a_{7}$, the (possibly not fully defined) target state. Generated trajectories consist of up to $n = 7$ phases of constant jerk input, resulting in bang-singular-bang trajectories.

\noindent Depending on:
\begin{compactitem}
    \item whether the trajectory shall be time-optimal or with defined total duration,
    \item which DoF is defined (undefined DoF are indicated as $\textit{NaN}$),
    \item which state limits are enforced (unlimited states are represented by setting the corresponding limit to $\textit{Inf}$),
    \item whether the trajectory starts with state limits already violated,   
\end{compactitem}
we define sets of 21 second-order conditions to make the system of differential equations solvable. For example, when only acceleration and position is defined, like in \reffig{fig:partially_defined}, acceleration and velocity limits are not $\textit{Inf}$, and the trajectory start is feasible, 20 different second-order condition combinations are possible. The optimal yellow trajectory in \reffig{fig:partially_defined}, for example, features only three constant jerk times. Thus, $t_{4} , ..., t_{7} := 0$.  Furthermore, $a_{2} := a_{max}$, $j_{1} := j_{max}$, $j_{2} := 0$, and $j_{3} := j_{min}$. In contrast, the red trajectory  features five jerk inputs. It is defined by the second-order conditions $t_{4} \land t_{5} := 0$, $a_{2} := a_{max}$, $a_{6} := a_{min}$, $j_{1} \land j_{7} := j_{max}$, $j_{2} \land j_{6} := 0$, $j_{3} := j_{min}$, and $t_{1} + ... + t_{7}:= T$.
After defining the set of possible second-order conditions, we solve the equations. For $t_{1}$ and $t_{3}$ of the yellow example trajectory, this yields
\begin{align}
t_{1} &= \frac{-a_{init} + a_{max}}{j_{max}}, \\
t_{3} &= \frac{-a_{max} + a_{target}}{j_{min}}.
\end{align}
The solution for $t_{2}$ is more complex and consists of $\approx 200$ computational operations.

All equations that solve $t_{1}, ..., t_{7}$ for a particular trajectory shape are calculated only once and stored in a database to be evaluated quickly during operation. We make no assumption on the value of any variable during solution so that all solutions are persistent. Since they neither change before nor during runtime, no precomputation is needed.

Trajectories do not only respect constraints during the trajectory time interval, but the target state also is defined such that future state constraints are not infringed. For example, when acceleration is an undefined DoF and the target velocity is close to the maximum allowed velocity, the chosen acceleration is not allowed to be positive. Although velocity as well as acceleration are inside the allowed bounds, a future trajectory would overshoot the maximum velocity during the deceleration phase. The maximum allowed target state velocity $v_{target}$ with a given acceleration is computed according to \refeq{eq:v_target} and the corresponding acceleration $a_{target}$ at a given velocity according to \refeq{eq:a_target}:
\begin{align}
  v_{target} &= \frac{a_{max}^2}{2 \cdot j_{max}} + v_{min},
  \label{eq:v_target}\\
  a_{target} &= \sqrt{2 \cdot j_{max} \cdot (v_{max} - v_{min})}.
  \label{eq:a_target}
\end{align}
As previously stated, trajectories can be time-optimal or have a fixed duration. If the duration is fixed, our method can be configured to maximize or minimize any undefined DoF. In \reffig{fig:partially_defined}, one can see that in comparison to the yellow trajectory, the additional time of \SI{2.5}{\second} is used to gain/loose velocity. This results in a target velocity of \SI{2.3}{\meter\per\second}, respective \SI{-0.17}{\meter\per\second}, in comparison to \SI{1.9}{\meter\per\second}. This behavior does not only work for the depicted example with undefined velocity, but also for all other combinations from \reftab{tab:permutations}.

\section{Obstacle Avoidance}
\label{sec:Obstacle_Avoidance}
A collision with an obstacle in n-dimensional space happens exactly when at one time all positions of all axes of the trajectory are simultaneously inside a section of a corresponding n-dimensional obstacle.
We assume the axes of the obstacles to be independent and axis-aligned. For two dimensions, this leads to rectangular obstacles. For three dimensions, this generates cuboid obstacles, and for higher dimensions hyperrectangles. Not only the position of the obstacle is defined in n-dimensions, but also its velocity, acceleration, etc.

\begin{figure}[t]
  \centering
  \begin{tikzpicture}[font=\sffamily,>={Stealth[inset=0pt,length=4pt,angle'=45]}]]

  \definecolor{red}{rgb}     {0.7,0.0,0.0}
  \definecolor{green}{rgb}   {0.0,0.7,0.0}
  \definecolor{blue}{rgb}    {0.0,0.0,0.7}
  \definecolor{grey}{rgb}    {0.5,0.5,0.5}

  \def\xscaling{3.5}
  \def\yscaling{0.7}
  
  \tikzset{content_node/.append style={black,text=black,fill=black!10!white,minimum size=1.5em,minimum width=2em,draw,align=center,rounded corners,scale=0.5}}
  \tikzset{label_node/.append   style={midway,align=center,scale=0.4}}
  \tikzset{group_node/.append   style={scale=0.6}}

  \draw[fill=black](0.0,0.1)circle(1pt);
 
  \node(Check_Velocity_Influence)[content_node] at (0.0*\xscaling,-0.5*\yscaling) {Check Velocity Influence};
  
  \node(Check_Signs)[content_node] at (-0.625*\xscaling,-1.0*\yscaling) {Check Signs};
  \node(Min_opt)[content_node,fill=green!30!white] at (-1.0*\xscaling,-2.0*\yscaling) {Min opt.};
  \node(Max_opt)[content_node,fill=green!30!white] at (-0.75*\xscaling,-2.0*\yscaling) {Max opt.};
  \node(Zero_1)[content_node,fill=green!30!white] at (-0.5*\xscaling,-2.0*\yscaling) {Zero};
  \node(Zero_2)[content_node,fill=green!30!white] at (-0.25*\xscaling,-2.0*\yscaling) {Zero};
   
  \node(Check_Configuration)[content_node] at (0.625*\xscaling,-1.0*\yscaling) {Check Configuration};
  \node(Segm12_opt)[content_node,fill=red!30!white] at (0.25*\xscaling-0.6,-2.0*\yscaling) {Segm. 1$\land$2 opt.};
  \node(Segm1_opt)[content_node,fill=red!30!white] at (0.5*\xscaling-0.15,-2.0*\yscaling) {Segm. 1 opt.};
  \node(Segm2_opt)[content_node,fill=red!30!white] at (0.75*\xscaling+0.15,-2.0*\yscaling) {Segm. 2 opt.};
  \node(Tradeoff)[content_node,fill=red!30!white] at (1.0*\xscaling+0.6,-2.0*\yscaling) {Tradeoff};

  \begin{scope}[on background layer]
    \draw[->,thick] (0.0,0.1) -- node[label_node,left] {} (Check_Velocity_Influence);
  
    \draw[->,thick] (Check_Velocity_Influence) -- node[label_node,above] {Single Velocity\\Influence\\(\refsec{sec:Single_Velocity_Influence})} (-0.625*\xscaling,-0.5*\yscaling) -- (Check_Signs);
    \draw[->,thick] (Check_Signs) -- (-1.0*\xscaling,-1.0*\yscaling) -- node[label_node,right, pos=0.58,xshift=0.1cm] {$+ +$} (Min_opt);
    \draw[->,thick] (Check_Signs) -- (-0.75*\xscaling,-1.0*\yscaling) -- node[label_node,left, pos=0.58,xshift=-0.1cm] {$- -$} (Max_opt);
    \draw[->,thick] (Check_Signs) -- (-0.5*\xscaling,-1.0*\yscaling) -- node[label_node,right, pos=0.58,xshift=0.1cm] {$+ -$} (Zero_1);
    \draw[->,thick] (Check_Signs) -- (-0.25*\xscaling,-1.0*\yscaling) -- node[label_node,left, pos=0.58,xshift=-0.1cm] {$- +$} (Zero_2);
  
    \draw[->,thick] (Check_Velocity_Influence) -- node[label_node,above] {Multiple Velocity\\Influence\\(\refsec{sec:Multiple_Velocity_Influence})} (0.625*\xscaling,-0.5*\yscaling) -- (Check_Configuration);
    \draw[->,thick] (Check_Configuration) -- (0.25*\xscaling-0.6,-1.0*\yscaling) -- node[label_node,right, pos=0.58,xshift=0.1cm] {$p_{7-2-8}$\\$p_{5-4-6}$} (Segm12_opt);
    \draw[->,thick] (Check_Configuration) -- (0.5*\xscaling-0.15,-1.0*\yscaling) -- node[label_node,left, pos=0.58,xshift=-0.1cm] {$p_{7-2-8}$\\$\cancel{p_{5-4-6}}$} (Segm1_opt);
    \draw[->,thick] (Check_Configuration) -- (0.75*\xscaling+0.15,-1.0*\yscaling) -- node[label_node,right, pos=0.58,xshift=0.1cm] {$\cancel{p_{7-2-8}}$\\$p_{5-4-6}$} (Segm2_opt);
    \draw[->,thick] (Check_Configuration) -- (1.0*\xscaling+0.6,-1.0*\yscaling) -- node[label_node,left, pos=0.58,xshift=-0.1cm] {$\cancel{p_{7-2-8}}$\\$\cancel{p_{5-4-6}}$} (Tradeoff);
   \end{scope}
   
\end{tikzpicture}~
  %\vspace{-0.5ex}
  \caption{\small{Structure of the obstacle avoidance approach.}}
  \label{fig:structure}
  \vspace{-2.5ex}
\end{figure}
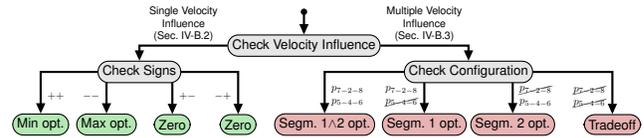

\subsection{Collision Detection}
\label{sec:Collision_Detection}
We assume the current state and the target state to be in free space. In order to efficiently detect collisions of the trajectory between both, we make use of the piecewise polynomial formulation of our trajectories. For every axis, the trajectory consists of a concatenation of up to seven polynomials with rank\,$\leq$\,3. Assuming that the obstacle motion can also be described with a concatenation of polynomials with rank\,$\leq$\,3, we can determine the time of the collision by calculation of zero crossings of the combined function. We then decide for every timestamp if the trajectory enters or leaves an obstacle. Given we found all entering and leaving times for all axes, we now merge the timestamps to search for times when all axes are simultaneously inside an obstacle. These timestamps indicate the entering and leaving of the n-dimensional trajectory into/out\,of the obstacle. Including the directional information stated above, we can detect the first entering which is the point of collision. Evaluating the polynomials at this timestamp gives the state-vector at the collision including velocity and acceleration. When we find a collision, we insert viastates into the trajectory such that the trajectory bypasses the obstacle. Finding the full viastate such that the trajectory is still time-optimal is not trivial and will be described in the following sections.

\subsection{Bound and Free Axes}
\label{sec:Bound_and_Free_Axes}
As stated above, a collision occurs only when all axes are simultaneously inside an obstacle. We insert a viastate into the trajectory to temporally disconnect this simultaneity. For this, we choose two axes that we want to temporally disconnect. We call these two axes ``bound axes''. We call all other axes (if there are more than two axes in total) ``free axes''. First, we describe how to derive the state vector for the two bound axes in \refsec{sec:State_Vector_for_Bound_Axes}. After this, we describe the derivation of the state vector of the free axes in \refsec{sec:State_Vector_for_Free_Axes}. The generated trajectory consist of a concatenation of up to 14 polynomials with rank\,$\leq$\,3 per axis. 

\begin{figure}[t]
  \centering
  \includegraphics[trim=190mm 25mm 190mm 35mm,clip,width=0.8\linewidth,height=0.83\linewidth]{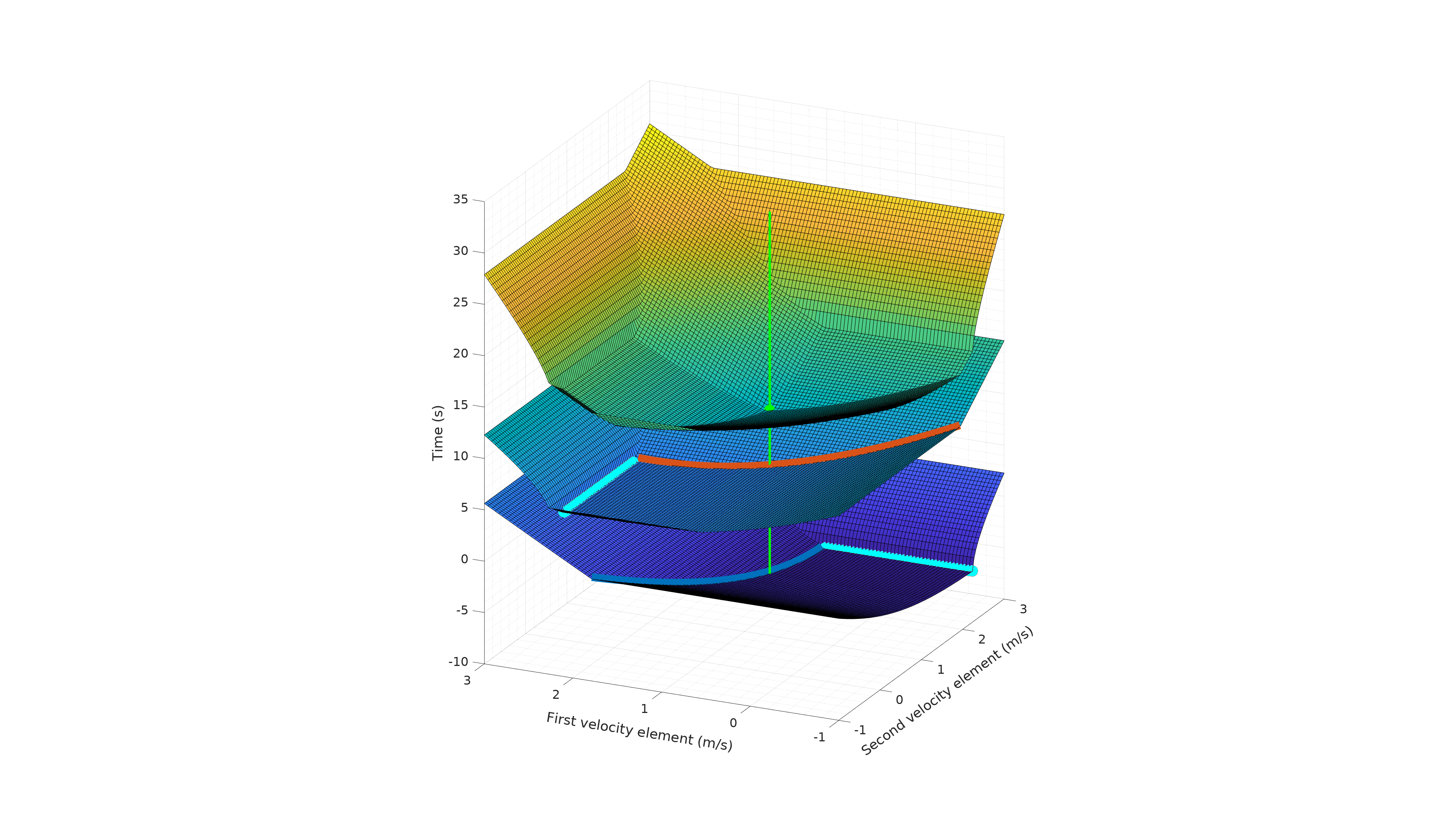}~
  %\vspace{-1ex}
  \caption{\small{Typical plot of the time needed for the first segment (middle), the second segment (bottom, shifted by 10s for better visibility) and the combined total trajectory (top). The time is shown in dependence of the velocity vector of the bound axes in the viastate. Local optima are marked with cyan dots. The global optimum is marked with a green dot and a vertical line. It can be seen that the global optimum lies at the intersection of one groove (marked blue and orange) of the individual segments. In \refsec{sec:Multiple_Velocity_Influence}, we detail our elaborated method to efficiently calculate these grooves and their intersection.}}
  \label{fig:3d_timing}
  \vspace{-2ex}
\end{figure}

\subsubsection{State Vector for Bound Axes}
\label{sec:State_Vector_for_Bound_Axes}

We first define the viastate position of the bound axes to be at one of the four corners of the obstacle and project the MAV radius + an additional margin outwards.
This determination is intuitive, since in order to find the fastest trajectory, one has to cut the corner as narrow as possible.

We now define the acceleration in these viastates to be zero. Setting the acceleration to the maximum/minimum in the viastates could gain a small amount of velocity, but if the state estimation of either the MAV or the obstacle is only slightly erroneous, the MAV is in a configuration that is close to the obstacle and possibly further accelerating towards the obstacle. By fixing the acceleration to zero, the MAV can react to these inaccuracies. So, a nonzero acceleration could be slightly faster (with massively increased collision risk), but under the assumption of zero acceleration, the generated evading trajectories are optimal.

Following this, we calculate the optimal viastate velocity of the two bound axes. \reffig{fig:3d_timing} shows a sampled map of the time needed to complete the first segment from start to the viastate (middle), the second segment from the viastate to the target state (bottom, shifted by 10s for better visibility) and the combined total time for a typical trajectory (top). Although trajectory sampling with our method is fast ($\approx$ \SI{519}{\micro\second} per trajectory), sampling the 20,000 trajectories ($100 \times 100$ sampling grid for each segment) took \SI{10.38}{\second} which makes it impossible to run at the control-frequency of the MAV. Besides the disadvantage of computational demands, this sampling-based method is comparatively inaccurate due to discretization. Although not viable for real-time calculation of the optimal velocity, the sampled map gives insight into the dependence of the timing from the two velocities and can be used to verify our real-time approach.
We exploit the characteristics of the shown surface plots to develop a more sophisticated method and speed up the process by multiple orders of magnitude.
Among others, we make use of the methods for partially defined target states we described in~\refsec{sec:Partially_Defined_Target_States}. The process works as follows:

\begin{figure}[t]
  \centering
  \includegraphics[trim=00mm 00mm 00mm 00mm,clip,width=0.8\linewidth]{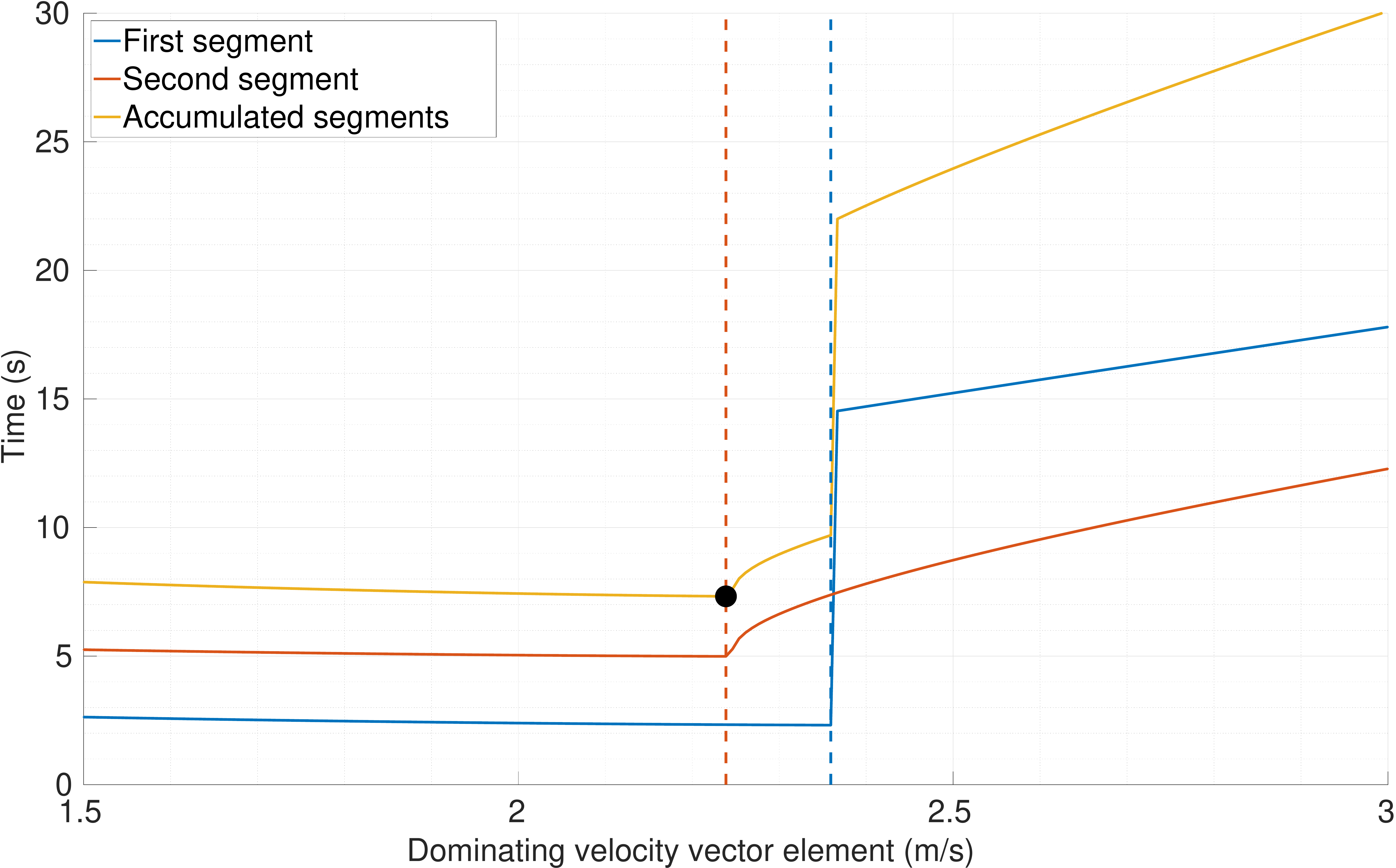}~
  %\vspace{-1ex}
  \caption{\small{Time needed for Segment~1 (blue), Segment~2 (orange), and total trajectory time (yellow) in dependence of a single velocity vector element. The other vector elements do not affect trajectory time. The optimum velocity for the individual segments is marked with a dashed line. The global optimum is marked with a black dot. The optimum is located at the optimal individual segment velocity that is closer to zero. This plot corresponds to the trajectory that goes over the obstacle in \reffig{fig:teaser}.}}
  \label{fig:trajectories_even}
  \vspace{-2ex}
\end{figure}

First, we use our method described in~\refsec{sec:Partially_Defined_Target_States} to find time-optimal trajectories from the current state to the position of the viastate of the bound axes with undefined velocity. As stated above, the velocity is chosen such that the trajectory is as fast as possible. We do the same backwards from the target state to the viastate position. This gives four velocities that would lead to optimal trajectories for both axes for the first and the second segment, respectively. We now compare which axis dominates the first and second segment. \reffig{fig:structure} shows the structure of the now following procedure.
If a single axis dominates both segments, we continue as stated in~\refsec{sec:Single_Velocity_Influence} (left in \reffig{fig:structure}). Otherwise, we continue as stated in ~\refsec{sec:Multiple_Velocity_Influence} (right in \reffig{fig:structure}).

\subsubsection{Single Velocity Influence}
\label{sec:Single_Velocity_Influence}
If only a single velocity component impacts the total trajectory time, again multiple cases can occur. If the signs of the velocities are different, we choose the velocity to be zero. While a positive velocity would speed up one segment of the trajectory, it would slow down the other segment even more. In this case, none of the individual segments is optimal, but the combination of both shows optimality.

If the sign of the velocities is equally positive, we use the minimum optimal velocity. This means that the segment with the smaller velocity is optimal resulting in the combination of both segments being optimal. See~\reffig{fig:trajectories_even} for clarification. Analogously, we choose the maximum optimal velocity if both signs are equally negative.

Since optimality is only depending on the velocity component we derived above, the other bound component can be chosen such that the total time of the trajectory does not change. To calculate the region of velocities that does not impact total trajectory time, we employ the approach we described in~\refsec{sec:Partially_Defined_Target_States}. We first determine the time the trajectory needs for the first and second segment with the computed optimal velocity individually. We then find the maximum and minimum velocity that is possible in this time for the second bound axis.  We superimpose the intervals and get an interval from that we can freely choose the velocity without compromising optimality. While we could, e.g., choose the midpoint of the interval, we choose the end of the interval that maximizes the curvature from the trajectory away from the obstacle.

\begin{figure}[t]
  \centering
  \includegraphics[trim=00mm 00mm 00mm 00mm,clip,width=1.0\linewidth]{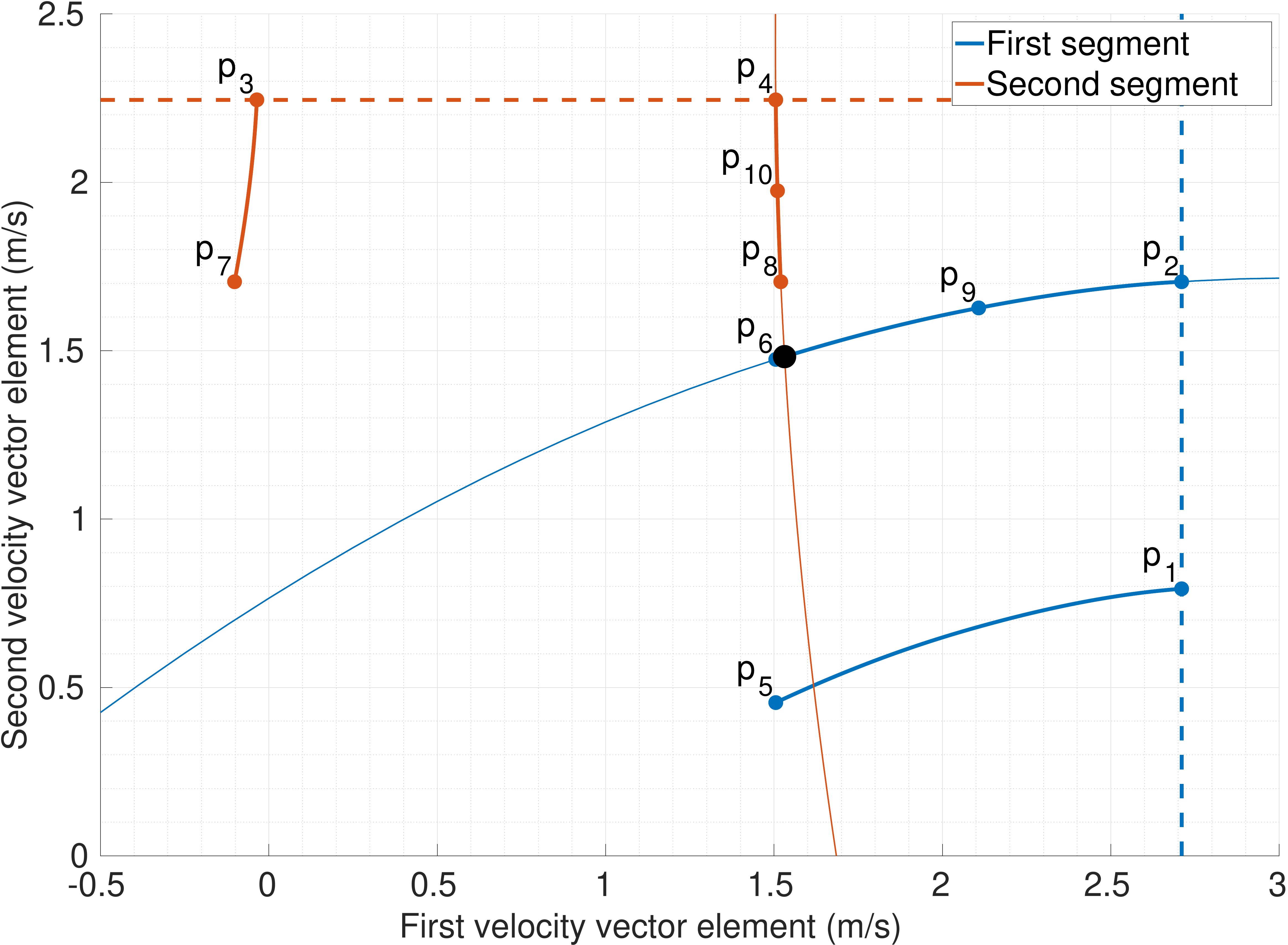}~
  %\vspace{-1ex}
  \caption{\small{Optimal tradeoff between two trajectory segments. For the first trajectory segment (blue), a viastate velocity of \SI{2.71}{\meter\per\second} (dashed line) is optimal. If this velocity would be chosen, the other velocity component could be freely chosen between $p_{1}$ and $p_{2}$ without changing the total trajectory time. The $p_{1\text{-}2}$ line corresponds to a local minimum in \reffig{fig:3d_timing}. For the second segment, a velocity of \SI{2.25}{\meter\per\second} would be optimal (orange dashed line). The other component is allowed to change between $p_{3}$ and $p_{4}$ (corresponding to the other local minimum \reffig{fig:3d_timing}). We first fit second order polynomials based on three dots per segment (thin lines). The lines $p_{2\text{-}9\text{-}6}$ and $p_{4\text{-}10\text{-}8}$ correspond to the blue and orange grooves in \reffig{fig:3d_timing}. We then calculate the intersection (black dot). This point represents the globally optimal velocity vector that is needed to generate time-optimal trajectories. This plot corresponds to the trajectory that goes left around the obstacle in \reffig{fig:teaser}.}}
  \label{fig:trajectories_uneven}
  \vspace{-2ex}
\end{figure}

\subsubsection{Multiple Velocity Influence}
\label{sec:Multiple_Velocity_Influence}
If both velocity components of the bound axes influence the total trajectory time, we employ the method illustrated in \reffig{fig:trajectories_uneven}. First we calculate the optimum velocities for both bound axes for both segments. We now determine which axis dominates the total time. The optimal velocity of the dominant axes are shown in \reffig{fig:trajectories_uneven}. We now determine the maximum and minimum velocities that are achievable in the time it takes to gain the optimal velocity ($p_{1}$, $p_{2}$, $p_{3}$, $p_{4}$) for both segments. Afterwards, we decide which point lies closer to the optimal counterpart of the other segment ($p_{2}$ respective $p_{4}$). This gives the direction of the limit.

\begin{figure}[t]
  \centering
  \includegraphics[trim=175mm 05mm 140mm 15mm,clip,width=0.8\linewidth]{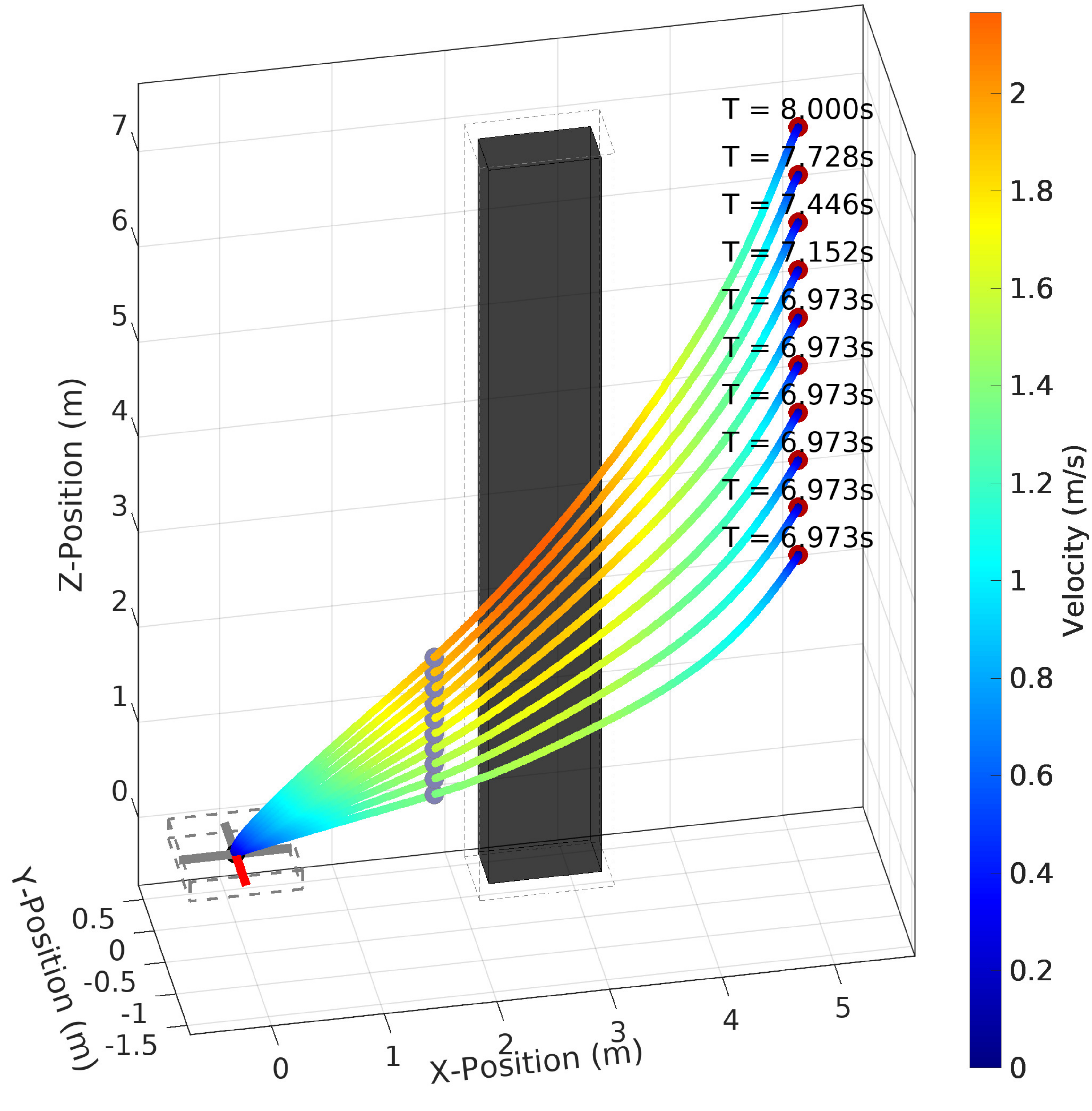}~
  \vspace{-1ex}
  \caption{\small{Trajectory stack with bound x- and y-axis and free z-axis. The transition between trajectories with dominating bound (bottom 6) and free (top 4) axes can be seen. The total trajectory duration does not change for the bottom 6 trajectories since the free z-axis is faster than both bound axes. The z-viastate adapts in position, velocity and acceleration. The free z-axis dominates the top 4 trajectories so that the viastate in the bound axes in turn adapts to the free axis.}}
  \label{fig:trajectories_free}
  \vspace{-4ex}
\end{figure}

After that, we determine $p_{5}$ and $p_{6}$, and $p_{7}$ and $p_{8}$ respectively. We do this by calculating the required time for Segment~1 to reach $p_{4}$ speed (\SI{1.51}{\meter\per\second}) and again the possible maximum and minimum velocity component. The velocity vector is the x-component from $p_{4}$ and the y-component of $p_{5}$ and $p_{6}$. The same is conducted for $p_{7}$ and $p_{8}$ with the y-component of $p_{2}$. Subsequently four cases can occur:\\
1) $p_{2}$ lies between $p_{7}$ and $p_{8}$ in x-direction and $p_{4}$ lies between $p_{5}$ and $p_{6}$ in y-direction. Both segments are optimal and the optimum velocity of both segments can be used.\\
2) $p_{2}$ lies between $p_{7}$ and $p_{8}$ in x-direction, but $p_{4}$ does not lie between $p_{5}$ and $p_{6}$ in y-direction. The first segment dominates the trajectory. The optimum lies at $p_{2}$ for both axes.\\
3) $p_{4}$ lies between $p_{5}$ and $p_{6}$ in y-direction, but $p_{2}$ does not lie between $p_{7}$ and $p_{8}$ in x-direction. The second segment dominates the trajectory. The optimum lies at $p_{4}$ for both axes.\\
4) Neither does $p_{2}$ lie between $p_{7}$ and $p_{8}$ in x-direction, nor does $p_{4}$ lie between $p_{5}$ and $p_{6}$ in y-direction.
Here, a tradeoff between the calculated velocities determines the optimum. This case is depicted in \reffig{fig:trajectories_uneven}. The tradeoff is calculated by first determining $p_{9}$ and $p_{10}$. The x-coordinate of $p_{9}$ lies in the middle of $p_{2}$ and $p_{6}$. The y-direction is again calculated by first determining the time needed to achieve $p_{9}$ speed. Afterwards, the maximum/minimum (depending on the direction of the limit) achievable velocity in this time is calculated. This gives the y-coordinate of $p_{9}$. Analogously, $p_{10}$ is derived.
We now approximate the optimal tradeoff for both bound axes separately by fitting a second order polynomial (\refeq{eq:axis_1} and \refeq{eq:axis_2}) through points $p_{2}$, $p_{9}$, and $p_{6}$, respectively $p_{4}$, $p_{10}$, and $p_{8}$ (thin lines). Subsequently, we now find the intersection of both polynomials by inserting \refeq{eq:axis_2} into \refeq{eq:axis_1}:
\begin{align}
  y &= a_{1} \cdot x^2 + b_{1} \cdot x + c_{1},
  \label{eq:axis_1}\\
  x &= a_{2} \cdot y^2 + b_{2} \cdot y + c_{2},
  \label{eq:axis_2}\\
  0 &= t^4 \cdot a_{1}^2 \cdot a_{2} 
  \label{eq:combined}\\
  & \quad + t^3 \cdot (2 \cdot a_{1} \cdot b_{1} \cdot a_{2}) \nonumber \\
  & \quad + t^2 \cdot (2 \cdot a_{1} \cdot c_{1} \cdot a_{2} + b_{1}^2 \cdot a_{2} + a_{1} \cdot b_{2}) \nonumber \\
  & \quad + t \cdot (2 \cdot b_{1} \cdot c_{1} \cdot a_{2} + b_{1} \cdot b_{2} - 1) \nonumber \\
  & \quad + c_{1}^2 \cdot a_{2} + c_{1} \cdot b_{2} + c_{2}. \nonumber
\end{align}
This results in the fourth-order polynomial in \refeq{eq:combined}. We do not simply equalize both equations since in the frame of the first axis the second axis is a transposed polynomial.
By finding the root of \refeq{eq:combined} that lies in the interval shown, we find the optimal tradeoff velocity vector marked with a black dot in \reffig{fig:trajectories_uneven}.

\subsubsection{State Vector for Free Axes}
\label{sec:State_Vector_for_Free_Axes}
After both 3-dimensional viastates of the two bound axes are calculated, we calculate the corresponding state for the remaining free axes.

For this, we first compute optimal trajectories for all free axes. We then compare the maximum free trajectory time with the bound trajectory time and choose the viastate of the free axes, depending on which one is larger.

If the bound axes dominate the total trajectory (bottom six trajectories in \reffig{fig:trajectories_free}), we synchronize all free axes to the total trajectory time of the bound axes by the methods presented in~\cite{beul2017icuas}. The derivatives in the viastate of the free axes are simply the states of the free axes at the time the bound axes pass the viastate.
If the free axes dominate the total trajectory, however, (top four trajectories in \reffig{fig:trajectories_free}) the free axis determines the viastate and we discard the viastate calculated in~\refsec{sec:State_Vector_for_Bound_Axes}. Instead, the viastate for the bound axes is the state when it passes the obstacle edge. Since the viastate for the bound axes is uncritical to guarantee time-optimality, it can vary as long as the free axis still dominates the total trajectory time. We choose the bound axis viastate such that the viastate is passed at a time that scales the trajectory segments.

\reffig{fig:trajectories_free} illustrates multiple trajectories with bound x/y axes and a varying free z-axis. It shows that the total trajectory time is constant as long as the bound axes dominate the trajectory. Only the top four trajectories are dominated by the free axis.

\section{Evaluation}
\label{sec:Evaluation}
We evaluate our method in simulation with arbitrarily placed obstacles. The dynamics of the MAV and the obstacle are simulated with MATLAB utilizing a realistic parameter set. A video showing avoidance of static and non-static obstacles can be found on our website\footref{ftn:Website}. Here, we also publish recorded datasets, tools, and parts of our pipeline.

We first evaluate our approach to partially defined target states. Based on this, we evaluate our obstacle avoidance pipeline. Since we conducted extensive flight tests with the predecessor of our software published in \eg \cite{ecmr2017_c3}, \cite{beul_2018_iros}, and \cite{ecmr2017_c1}, the practical applicability of the method to real MAVs ranging from \SI{3.0}{\kilo\gram} to \SI{11.2}{\kilo\gram} has already been shown.

\subsection{Partially Defined Target States}
\label{sec:eval_Partially_Defined_Target_States}
First, we evaluate the reliability of our method by fuzz testing our approach. Therefore, random data including $\textit{NaN}$ for the target state and $\textit{Inf}$ for the state limits are generated. Our method successfully found solutions for over 10,000,000 consecutive trajectories.% so that we assume this feature to be working correctly.

\subsection{Obstacle Avoidance}
\label{sec:eval_Obstacle_Avoidance}
In order to test the soundness of our approach, obstacle avoidance was tested in over 100 different scenarios with changing obstacle configurations.

\begin{figure}[t]
  \centering
  \includegraphics[trim=00mm 00mm 00mm 00mm,clip,width=1.0\linewidth]{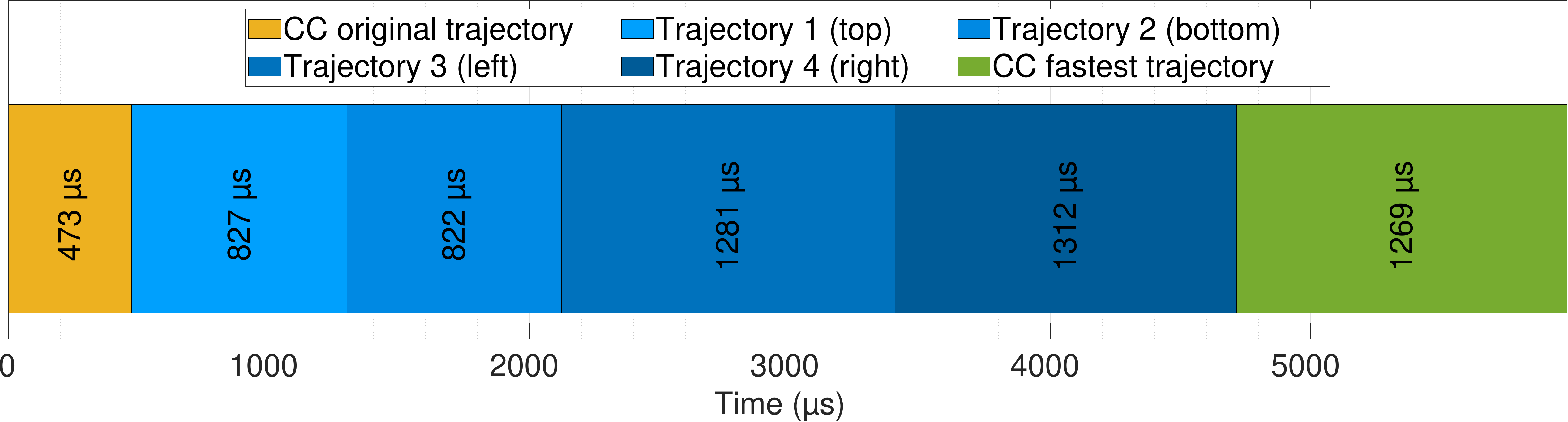}~
  \vspace{-1ex}
  \caption{\small{Results of the timing experiment for the three-dimensional trajectories shown in \reffig{fig:teaser}. The total computation time of \SI{5.98}{\milli\second} is mostly needed for generating the individual evading trajectories. Collision checking (CC) only takes a quarter of the time.}}
  \label{fig:timing}
  \vspace{-3ex}
\end{figure}

We measure the time, our obstacle avoidance method requires on a single core of a Intel Core i7-4710MQ processor for the scenario shown in \reffig{fig:teaser}. In \reffig{fig:timing}, we show the results of our profiling for the experiment. It can be seen that our whole pipeline computes in less than \SI{6}{\milli\second}, including collision checking. Since the four individual trajectories are independent of each other, this process can be easily parallelized.

\begin{figure*}[t]
  \begin{center}
  \begin{subfigure}[b]{0.392\textwidth}
      \includegraphics[trim=75mm 48mm 61mm 66mm,clip,width=1.0\linewidth]{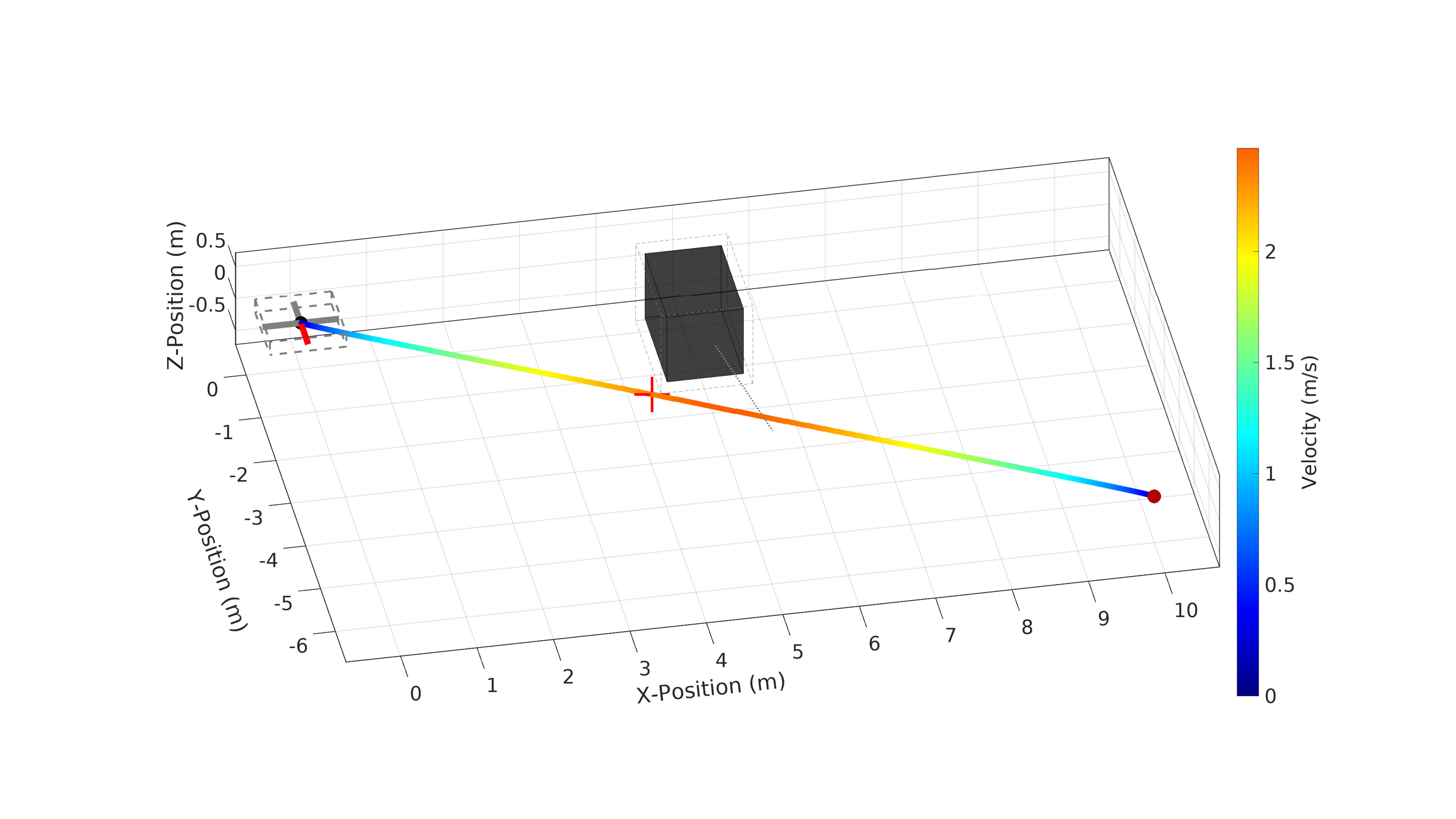}~
      \vspace{-1ex}
      \caption{}
  \end{subfigure}
  \hspace{0.1ex}
  \begin{subfigure}[b]{0.392\textwidth}
      \includegraphics[trim=75mm 48mm 61mm 66mm,clip,width=1.0\linewidth]{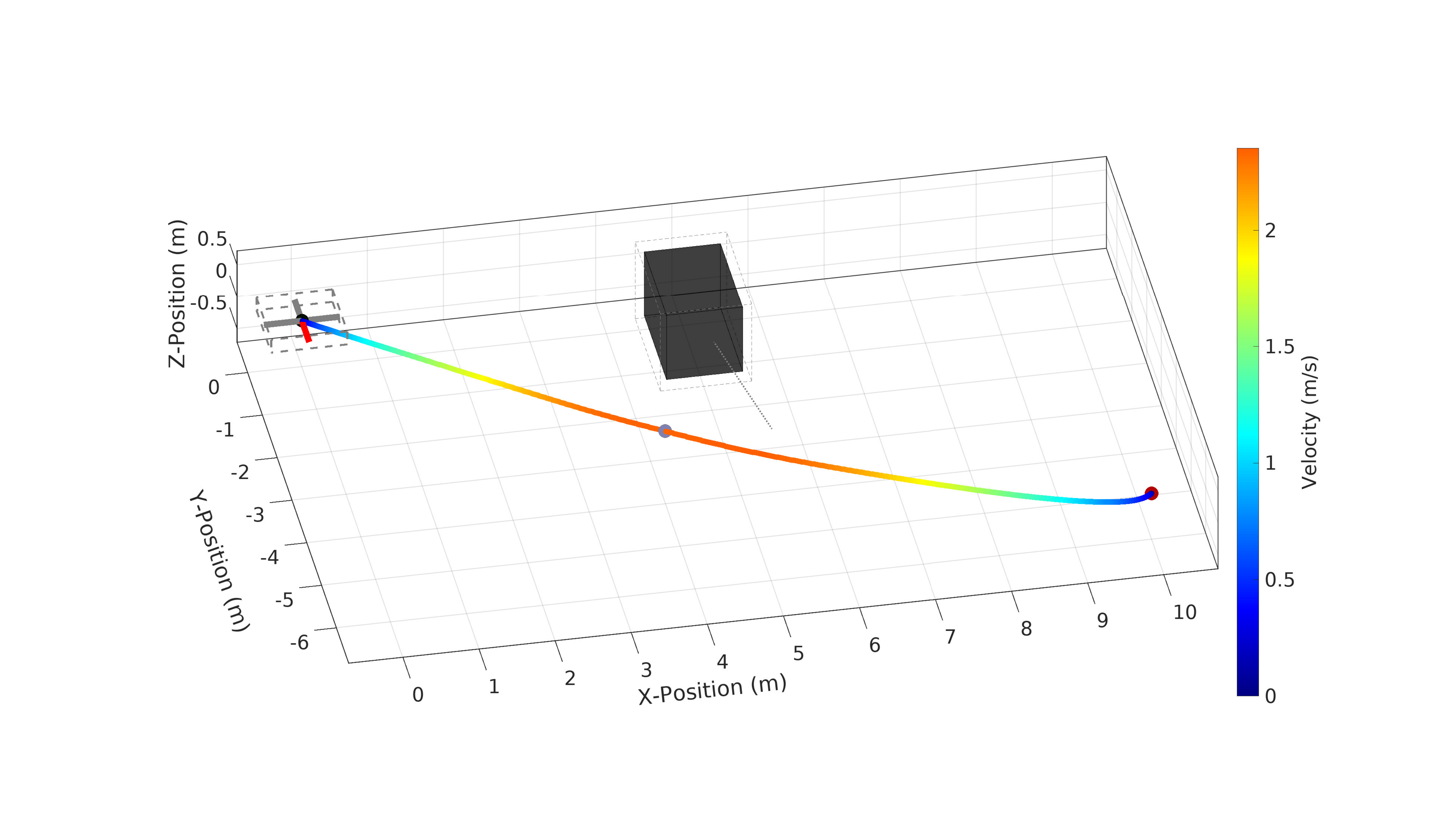}~
      \vspace{-1ex}
      \caption{}
  \end{subfigure}
  %\centering
  \begin{subfigure}[b]{0.392\textwidth}
      \includegraphics[trim=75mm 48mm 61mm 66mm,clip,width=1.0\linewidth]{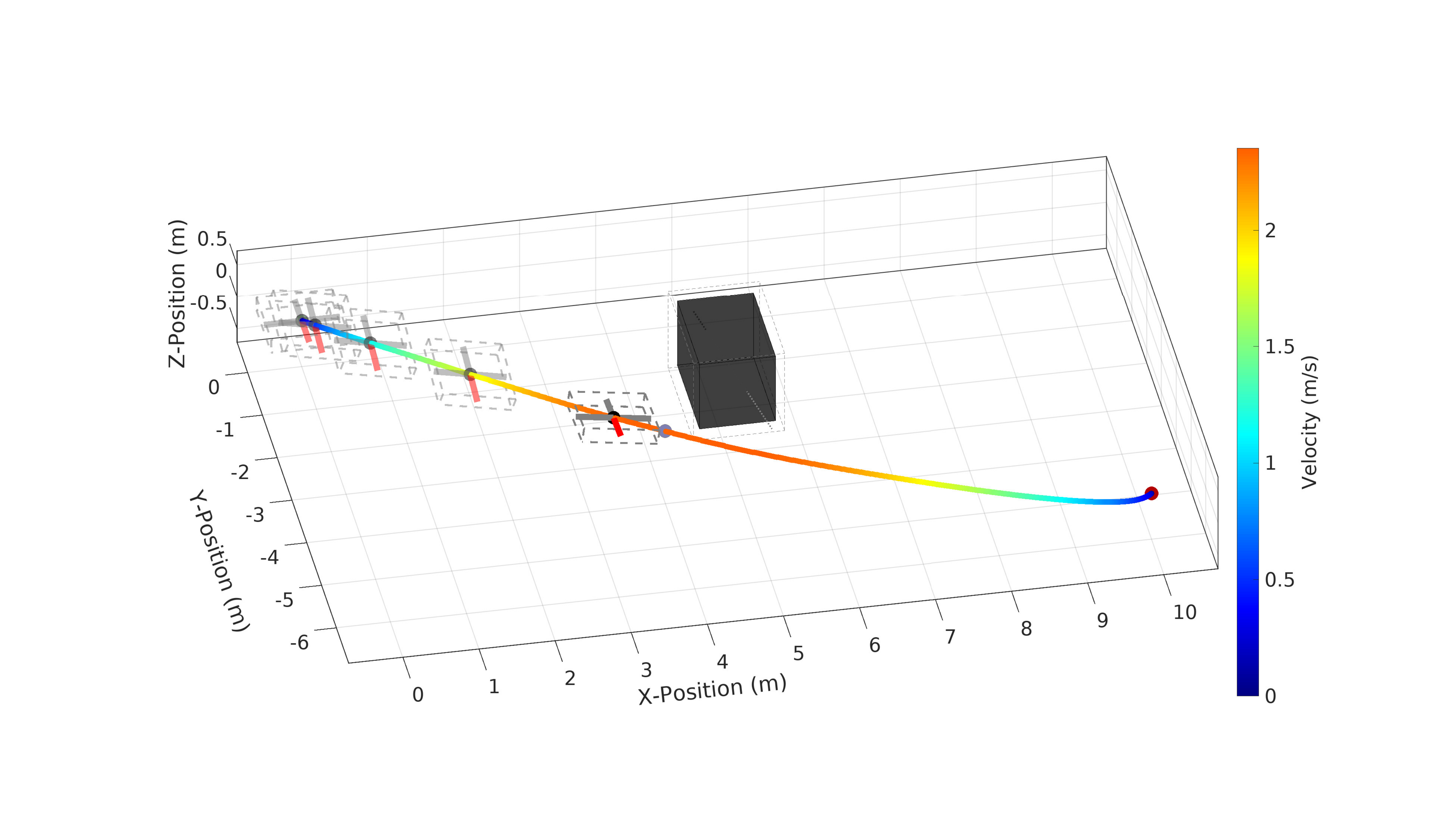}~
      \vspace{-1ex}
      \caption{}
  \end{subfigure}
  \hspace{0.1ex}
  \begin{subfigure}[b]{0.392\textwidth}
      \includegraphics[trim=75mm 48mm 61mm 66mm,clip,width=1.0\linewidth]{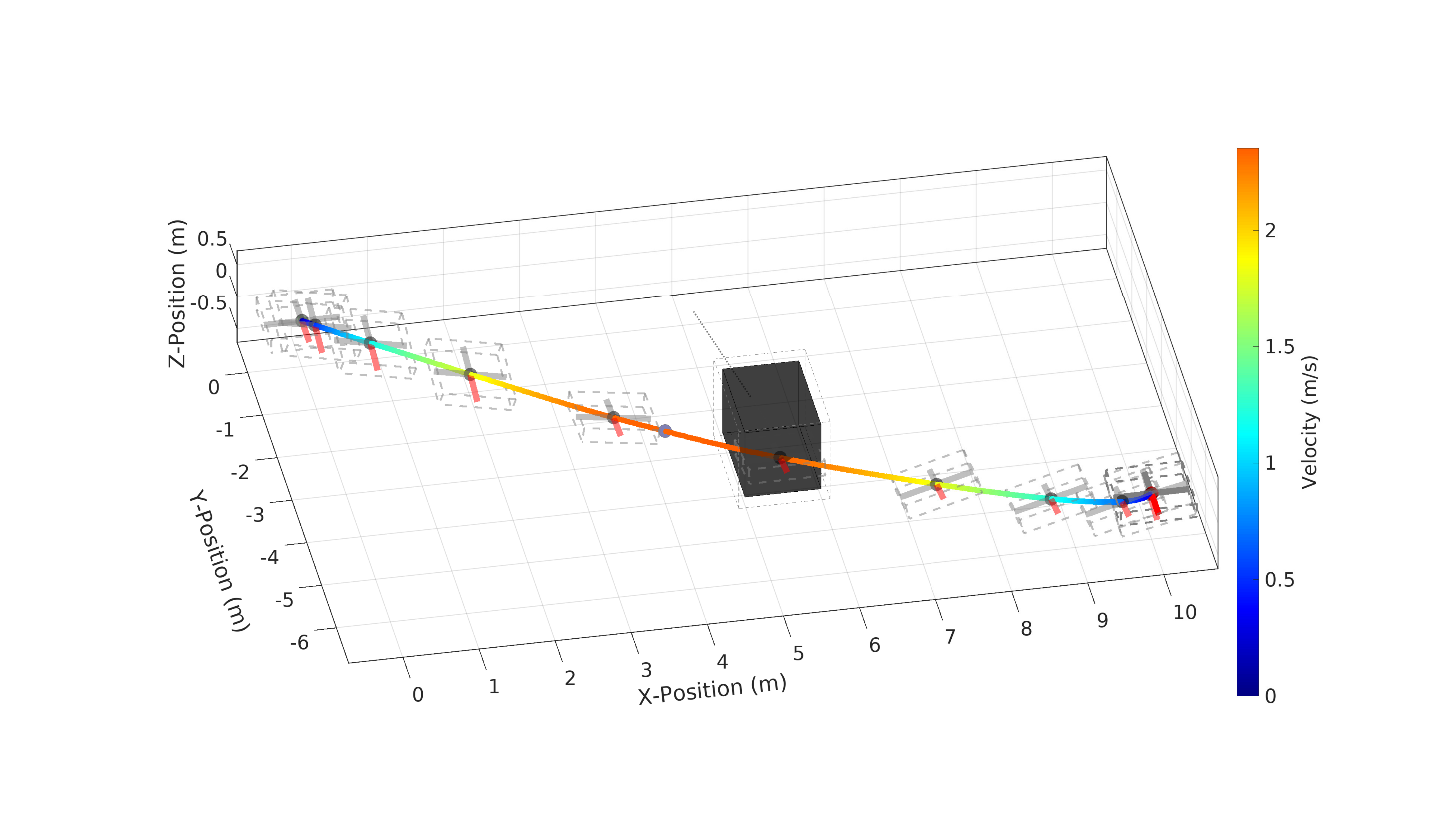}~
      \vspace{-1ex}
      \caption{}
  \end{subfigure}
  \end{center}
  \vspace{-1ex}
  \caption{\small{Avoiding a moving obstacle. (a) Our method detects a future collision at the position marked with the red cross of the original trajectory and the moving obstacle trajectory (marked with the gray dotted line). With a static obstacle no collision would occur, but since our pipeline explicitly considers the movement, the collision is reliably detected. (b) It now generates an optimal viastate (gray) including an optimal velocity vector that guides the MAV around the obstacle incorporating the movement of the object. (c) The trajectory is executed. (d) The MAV successfully avoided the obstacle. The trajectory only took  \SI{9.52}{\second} vs. the original \SI{9.45}{\second}.}}
  \label{fig:moving_obstacle}
  \vspace{-3ex}
\end{figure*}

We further extend our evaluation to a constantly moving obstacle depicted in \reffig{fig:moving_obstacle}. We observe that the movement has no measurable impact on the computation time.
We also examine the dependence of the runtime to the dimensionality of the planning problem. Since our method always temporally disconnects two axes per instance, the number of possible evading trajectories $p$ regarding n axes scales with
\begin{equation}
 p = \frac{n!}{(n-2)!}.
\label{eq:scaling}
\end{equation}

Experiments with four axes (x, y, z, yaw) also confirm this. Here, obstacles are 4-dimensional and problems like ``Do not look in direction X while being in the volume Y.'' can be represented. Depending on the configuration of the start- and target conditions, one can further omit several bound axis combinations. Thus, our experiments only show four trajectories instead of six. In \reffig{fig:teaser}, the collision of the original trajectory happens in the y/z plane. Thus, the bound axis-combinations are x/y (left/right) and x/z (top/bottom). A y/z combination makes no sense since the main axis of movement is in x-direction.

Before executing the fastest trajectory of the set, it is checked for collisions with all obstacles (green bar in \reffig{fig:timing}). If it collides with an obstacle, the next bests trajectory is chosen. This process is repeated until the fastest trajectory without collision is found. So, if for example another obstacle was standing on top of the obstacle in \reffig{fig:teaser}, the next best trajectory below the obstacle would be selected.

Please note that our method does \textit{not} explicitly consider uncertainty in state estimation of MAV and obstacle and assumes a constant motion of the obstacle. However, due to the fast computation time, this can be compensated by a high replanning rate. The high replanning rate also promotes the performance when obstacles are not known in advance but are perceived abruptly and with uncertain velocity estimates. Regarding uncertainty, the position margin of the obstacle can be adjusted depending on the used sensor. Also due to the obstacle representation by two independent borders, obstacles can also grow or shrink over time representing growing uncertainty over time.

\section{Conclusion}
\label{sec:Conclusion}
In this paper, we propose a novel method to plan smooth, time-optimal obstacle avoiding trajectories which is realized through viastate insertion. Additional viastates are inserted into the trajectory and guide the trajectory around the obstacle. Finding the optimal position derivatives such that the trajectory is still time-optimal is accomplished by a method that requires only minimal computational effort.

Our evaluation shows that due to the fast runtime of our method, the approach can be used in real-time to avoid suddenly perceived obstacles in the flight path of the MAV. Our approach works with an arbitrary number of dimensions and with constantly moving obstacles.

We demonstrated the method in simulation and fuzz-tested components of the system several million times to show the robustness of the approach.

\IEEEtriggeratref{7}

\bibliographystyle{IEEEtran}
\bibliography{literature_references}

\end{document}